\documentclass[sigconf,screen,nonacm]{acmart}
\AtBeginDocument{%
  }
\setcopyright{none}
\renewcommand\footnotetextcopyrightpermission[1]{}
\usepackage{amsmath}  
\usepackage{graphicx}                  
\usepackage{booktabs,multirow,array}   
\usepackage{xcolor,colortbl}           
\usepackage{algorithm}                 
\usepackage{algorithmicx,algpseudocode}
\usepackage{url,verbatim}              
\newcommand{\best}[1]{\cellcolor{red!10}{#1}}
\newcommand{\second}[1]{\cellcolor{blue!10}{#1}}
\definecolor{mygrey}{cmyk}{0,0,0,0.5}

\begin{document}

\pagestyle{plain}

\title{Fast Model-guided Instance-wise Adaptation Framework for Real-world Pansharpening with Fidelity Constraints}

\author{Zhiqi Yang}
\email{zhiqi.Young@outlook.com}
\authornote{These authors contributed equally to this work.}
\affiliation{%
  \institution{University of Electronic Science and Technology of China}
  \city{Chengdu}
  \country{China}}

\author{Jin-Liang Xiao}
\email{jinliang\_xiao@163.com}
\authornotemark[1]
\affiliation{%
  \institution{University of Electronic Science and Technology of China}
  \city{Chengdu}
  \country{China}}

\author{Shan Yin}
\email{yins@std.uestc.edu.cn}
\affiliation{%
  \institution{University of Electronic Science and Technology of China}
  \city{Chengdu}
  \country{China}}

\author{Liang-Jian Deng}
\email{liangjian.deng@uestc.edu.cn}
\affiliation{%
  \institution{University of Electronic Science and Technology of China}
  \city{Chengdu}
  \country{China}}

\author{Gemine Vivone}
\email{gemine.vivone@imaa.cnr.it}
\affiliation{%
  \institution{National Research Council, Institute of Methodologies for Environmental Analysis (CNR-IMAA)}
  \city{Tito}
  \country{Italy}}

\begin{abstract}
Pansharpening aims to generate high-resolution multispectral (HRMS) images by fusing low-resolution multispectral (LRMS) and high-resolution panchromatic (PAN) images, with the goal of preserving both spectral and spatial information. Although deep learning (DL)-based pansharpening methods have achieved impressive performance, they typically come with high training costs and require large amounts of training data. Moreover, their performance often degrades significantly when the test data distribution differs from that of the training data, posing challenges for generalization ability. 
In recent years, zero-shot methods, which train on single test PAN/LRMS pair, have emerged as a promising alternative with strong generalization. However, most existing zero-shot methods suffer from limited fusion performance, incur significant computational overhead, and slow convergence, hindering their practical deployment. To address these challenges, we propose FMG-Pan, a fast and generalizable model-guided Instance-wise adaptation framework for real-world pansharpening, designed to achieve both \emph{cross-sensor generality} and \emph{rapid training-inference processing}. This flexible framework can incorporate any pre-trained model as guidance and fully leverages its representational capacity to guide a relatively lightweight adaptive network through joint optimization under spectral and physical fidelity constraints. In particular, we design a new physical fidelity term to enhance spatial detail preservation.Extensive experiments on multiple real-world datasets, including both intra-sensor and cross-sensor scenarios, demonstrate that FMG-Pan achieves state-of-the-art performance. Specifically, on the WorldView-3 dataset, it completes the training-inference process for a $\text{512} \times \text{512} \times \text{8}$ image within 3 seconds on an RTX 3090 GPU, which is markedly faster than existing zero-shot methods, making it highly suitable for deployment.
\end{abstract}


\maketitle

\section{Introduction}
\begin{figure}[ht]
    \centering
    \includegraphics[width=1\linewidth]{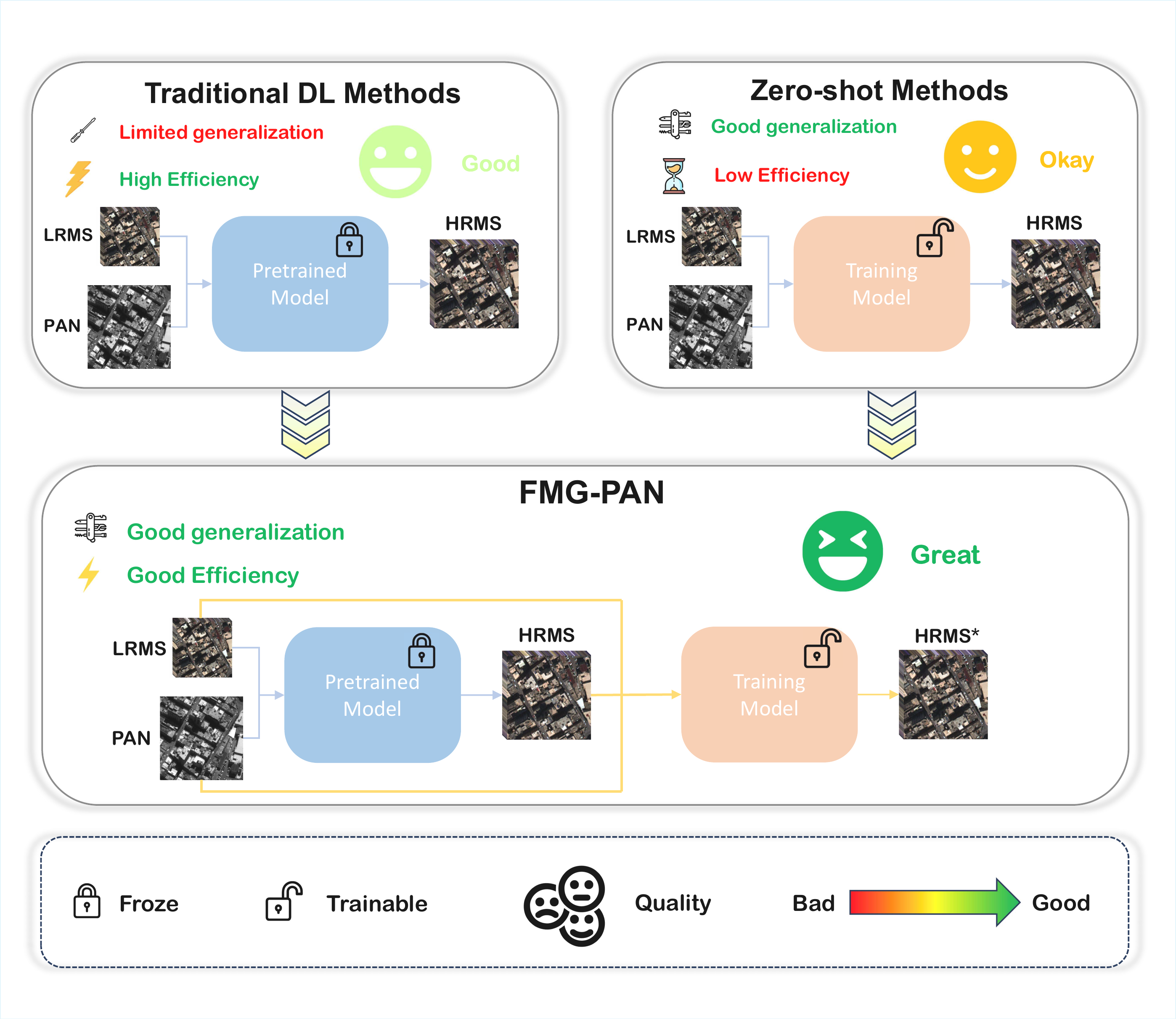}
    \caption{Comparison among traditional deep learning methods, zero-shot methods, and the proposed FMG-Pan framework in terms of performance, efficiency, and generalization. FMG-Pan achieves a better trade-off by leveraging pretrained models for fast, image-specific adaptation without modifying model architecture or requiring external training data.}
    \label{fig:enter-label}
\end{figure}
High-resolution remote sensing imagery is crucial for applications such as urban planning, precision agriculture, and environmental monitoring~\citep{vivone2020new, deng2022machine, vivone2024deep}. However, due to hardware limitations, satellite sensors cannot simultaneously acquire imagery with both high spectral and spatial resolution. Consequently, modern satellites (e.g., WorldView series) typically capture two complementary types of data: low-resolution multispectral (LRMS) images retaining spectral content and high-resolution panchromatic (PAN) images providing spatial detail. Pansharpening aims to fuse the LRMS and PAN images into a high-resolution multispectral (HRMS) image, combining spectral fidelity and spatial sharpness. This task is fundamental in remote sensing image processing, enabling high-quality data acquisition overcoming the limitations of the actual hardware devices in capturing images with both high spectral and spatial resolutions.

Over the past decades, researchers have proposed a variety of pansharpening methods, which can be broadly classified into four categories: (i) component substitution (CS), (ii) multi-resolution analysis (MRA), (iii) variational optimization (VO), and (iv) deep learning (DL)-based approaches. The first three are traditional model-driven methods based on explicit mathematical formulations: CS methods ~\citep{kwarteng1989extracting,carper1990use}, inject PAN details by replacing intensity components of the LRMS image; MRA methods ~\citep{liu2000smoothing,MTF-GLP-FS} employ pyramid or wavelet decomposition to extract multi-scale spatial information from the PAN image. VO methods\citep{zhu2012sparse,deng2019fusion}, treat HRMS generation as an optimization problem with carefully designed fidelity constraints. These methods are computationally efficient and require only a single PAN-LRMS input pair, without external training~\citep{xiao2023variational}. However, they often rely on simplified linear assumptions (e.g., linear correlation between PAN and HRMS images), which do not hold in complex or cross-sensor scenarios, leading to spectral or spatial distortions.

In recent years, deep learning methods have significantly advanced pansharpening by leveraging powerful representation capabilities of neural networks~\citep{masi2016pansharpening,yang2017pannet,DICNN,jin2022lagconv,peng2024fusionmamba,wu2025fully,cao2024diffusion,wang2024general,caotpami2025}. Supervised CNN-based models (e.g., PNN~\citep{masi2016pansharpening}, PanNet~\citep{yang2017pannet}) learn nonlinear mappings between PAN and LRMS through end-to-end training. Later works introduced attention mechanisms, Transformer architectures, and diffusion models to further improve fusion quality~\citep{zhong2024ssdiff, deng2023psrt, xiao2025hyperspectral}. Trained networks generally exhibit strong representational capacity and fast inference. However, deep learning approaches face three major challenges in practical deployment: (1) most models are trained on simulated downsampled data but \textit{tested on full resolution images}, resulting in resolution mismatch and performance drop; (2) they\textit{ heavily depend on large-scale training datasets}, leading to \textit{high training costs}; (3) They exhibit \textit{poor cross-domain generalization}, with performance degrading significantly when the test data distribution differs from the training set, even falling behind traditional methods. Although unsupervised methods reduce label dependence to some extent, they often rely on idealized degradation models and still require multiple samples to ensure robustness.

To address the above limitations, zero-shot or single-image training strategies have been proposed~\citep{PsDip, ZsPan, wang2024zero}. These approaches eliminate external data requirements by training directly on the target image via customized self-supervised loss functions, ensuring consistency between training and testing. This image-specific self-supervision enhances real-world performance without relying on additional datasets. However, such methods are often structurally complex, require precise degradation modeling, and involve time-consuming optimization on each image, which limits their practical use.

We propose a fast, model-guided pansharpening framework with strong generalization and plug-and-play adaptability. Different from conventional training-based methods, we treat the pretrained model as a black-box module and utilize its single-image output to guide a relatively lightweight adaptive network during image-specific training. With carefully designed spectral and physical fidelity losses, the adaptive network preserves prior structural information from the pretrained model while correcting potential distortions, achieving enhanced performance on the source dataset and improved generalization across sensors. Importantly, our method requires no access to the internal structure of the pretrained model and can be flexibly combined with existing models to support deployment across diverse sensors, resolutions, and domains with low cost and high adaptability. Fig.~\ref{fig:enter-label} compares traditional deep learning methods, zero-shot strategies, and the proposed FMG-Pan framework in terms of performance, efficiency, and generalization.

Our contributions are summarized as follows:
\begin{enumerate}
    \item We propose a model-guided framework (FMG-Pan) that addresses the challenges of quality, efficiency, and cross-sensor adaptability in real-world pansharpening, overcoming the low efficiency of zero-shot methods and the weak generalization of traditional DL methods.
    \item We introduce a novel physical fidelity loss that effectively utilizes input images information to further enhance the spatial and spectral quality of the fused HRMS image.
    \item Extensive experiments demonstrate that our method achieves state-of-the-art performance on multiple real-world remote sensing datasets within seconds, exhibiting strong adaptability and deployment efficiency. In particular, we achieved best results on WorldView-3 data completing the training-inference process within 3 seconds and on the WorldView-2 data within 10 seconds.
\end{enumerate}

The remainder of this paper is organized as follows. Section~\ref{RELATED} reviews related works and presents our motivations. Section~\ref{method} introduces the proposed framework. Section~\ref{experiments} reports experimental results on both intra-sensor and cross-sensor datasets, followed by ablation studies and discussions. Finally, Section~\ref{conclusion} concludes the paper.

\section{Related Works and Motivations}
\label{RELATED}
\subsection{Deep Learning-based Pansharpening}
Deep learning, with its superior feature representation and nonlinear modeling capabilities, has become the mainstream approach for pansharpening. From early methods, such as PNN~\citep{masi2016pansharpening}, to DiCNN~\citep{DICNN} and LAGNet~\citep{jin2022lagconv}, CNN-based models have achieved remarkable performance in spatial-spectral fusion. However, due to the limited receptive field of convolutional kernels, these methods often fail to capture global dependencies, leading to noticeable spatial distortions.

To overcome these limitations, Transformer-based architectures (e.g., Panformer~\citep{Panformer} and WFANet~\citep{WFANet}) have been proposed, which leverage attention mechanisms to model long-range spatial-spectral relationships and improve global consistency. Nonetheless, Transformers typically incur high computational costs and training complexity, making them difficult to deploy in resource-constrained scenarios. More recently, diffusion models have also been explored for pansharpening, showing potential in preserving spectral and spatial details~\citep{zhong2024ssdiff, pandiff}. However, most of these methods are trained on downsampled data and ignore full resolution information, which limits their applicability in real-world scenarios~\citep{9745494} and often leads to significant performance degradation in cross-sensor settings. In summary, although conventional deep learning methods can produce high-quality fusion results, they still suffer from limitations in training cost, cross-domain adaptability, and inference on real-world images.

\subsection{Traditional and Single-image Deep Pansharpening Methods}
Before the widespread adoption of deep learning, traditional model-driven approaches dominated pansharpening research, including component substitution methods (e.g., IHS~\citep{carper1990use}, PCA~\citep{kwarteng1989extracting}), multi-resolution analysis methods (e.g., Laplacian pyramid~\citep{Laplacian}, wavelet transform~\citep{wavelet-transform}), and variational optimization methods~\citep{zhu2012sparse,vo2,vo3}. These approaches offer strong interpretability, high computational efficiency, and independence from external training data. However, due to their reliance on overly simplified linear assumptions, they are prone to spectral/spatial distortions in complex scenarios leading to a performance reduction.

In recent years, researchers have introduced deep learning into single-image training strategies, which require no external training datasets and perform self-supervised optimization directly on the single target image. For example, ZS-Pan~\citep{ZsPan} employs a two-stage self-supervised process that combines degradation modeling with reconstruction, while PsDip~\citep{PsDip} embed CNNs into variational frameworks to achieve strong performance without external training. These methods exhibit inherent robustness in cross-domain tasks but face challenges such as complex loss function design and low inference efficiency, with inference speeds often significantly slower than conventional deep learning methods.



To solve this problem, we propose the FMG-Pan framework. It combines the strong representation ability of pretrained models with the adaptability of zero-shot methods. By using the pretrained output as guidance, our adaptive network converges faster, while spectral and physical fidelity losses help to reduce distortions. This results in a lightweight model-guided framework that preserves structure, achieves efficient inference, and shows strong generalization in real-world pansharpening tasks.

\section{Proposed Method}
\label{method}
\subsection{Notations}
The notations used throughout this paper are summarized as follows. Let $\mathbf{P} \in \mathbb{R}^{H \times W}$ denote the high-resolution PAN image, where $H$ and $W$ represent the height and width, respectively. The LRMS image is denoted by $\mathcal{Y} \in \mathbb{R}^{h \times w \times c}$, with $h$, $w$, and $c$ denoting the height, width, and number of spectral bands. The desired HRMS image is represented as $\mathcal{X} \in \mathbb{R}^{H \times W \times c}$. The reduced-resolution PAN image, obtained by downsampling the original PAN image, is denoted by $\widetilde{\mathbf{P}} \in \mathbb{R}^{h \times w}$, while the upsampled LRMS image is expressed as $\widehat{\mathcal{Y}} \in \mathbb{R}^{H \times W \times c}$. Finally, we define $r = \tfrac{H}{h} = \tfrac{W}{w}$ as the resolution scale factor between PAN and MS images.

\begin{figure*}[t]
  \centering
  \includegraphics[width=0.95\textwidth]{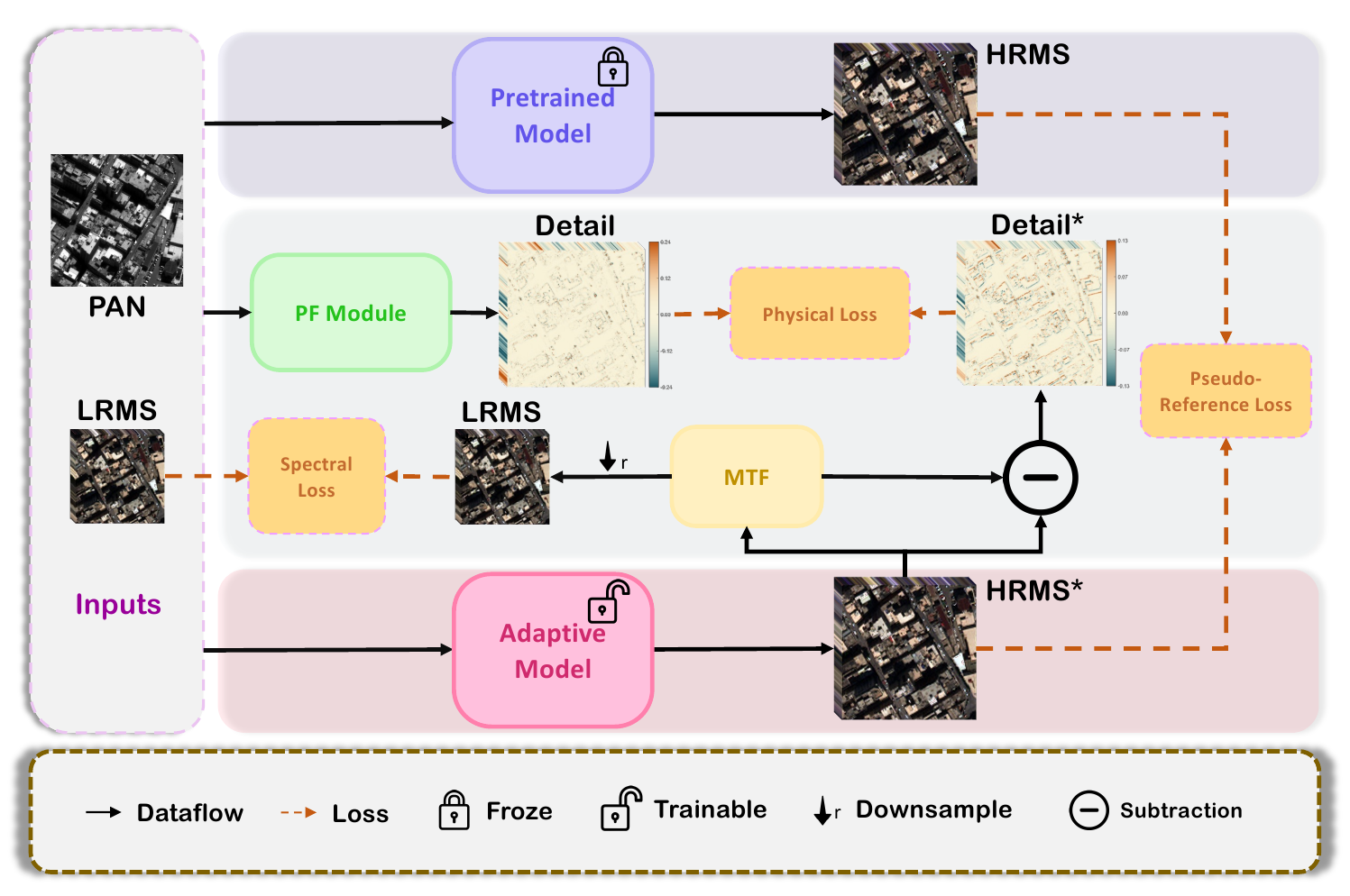}
  \caption{Overall framework of the proposed FMG-Pan framework. The pretrained model and the PF module are first executed independently. Subsequently, the parameters of the adaptive model are optimized using three core loss functions: pseudo-reference loss, spectral loss, and physical loss. In the reasoning stage, the final HRMS image is generated by applying the trained adaptive model to the input PAN and LRMS images.}
  \label{overall_framework}
\end{figure*}
\subsection{Overall Framework}
The overall framework of our proposed method, FMG-Pan, is illustrated in Fig.~\ref{overall_framework}. Given an LRMS image $\mathcal{Y} \in \mathbb{R}^{h \times w \times c}$ and a high-resolution PAN image $\mathbf{P} \in \mathbb{R}^{H \times W}$, the framework aims to generate a high-resolution multispectral (HRMS*) image $\mathcal{X}^{*} \in \mathbb{R}^{H \times W \times c}$ that simultaneously preserves spectral fidelity and spatial details.

The framework consists of four core components: a pretrained model, an adaptive model, a spectral fidelity module, and a physical fidelity module (PF module). At the beginning, the input images $\mathcal{Y}$ and $\mathbf{P}$ are fed into both the pretrained model and the PF module. The pretrained model, with fixed parameters, generates a pseudo-reference HRMS image, denoted as $\mathcal{X}$, serving as a soft label. Simultaneously, the PF module extracts spatial detail features from the PAN image, forming a detail map.

The adaptive model is then trained in a self-supervised manner, guided by three loss functions: the pseudo-reference loss, the spectral fidelity loss, and the physical fidelity loss. 

After the adaptive model converges, the trained adaptive model is directly applied to the input image pair to produce the final HRMS output $\mathcal{X}^{*}$.

It is worth noting that we adopt modulation transfer function (MTF)-matched filters to blur PAN and MS images. These filters approximate the MTF of the MS sensor, usually Gaussian-shaped, with the standard deviation set through the Nyquist frequency gain (typically from sensor manufacturers). The use of such filters is common in pansharpening, see~\citep{aiazzi2006mtf,vivone2014critical} for details.

\begin{figure*}[t]
  \centering
  \includegraphics[width=0.95\textwidth]{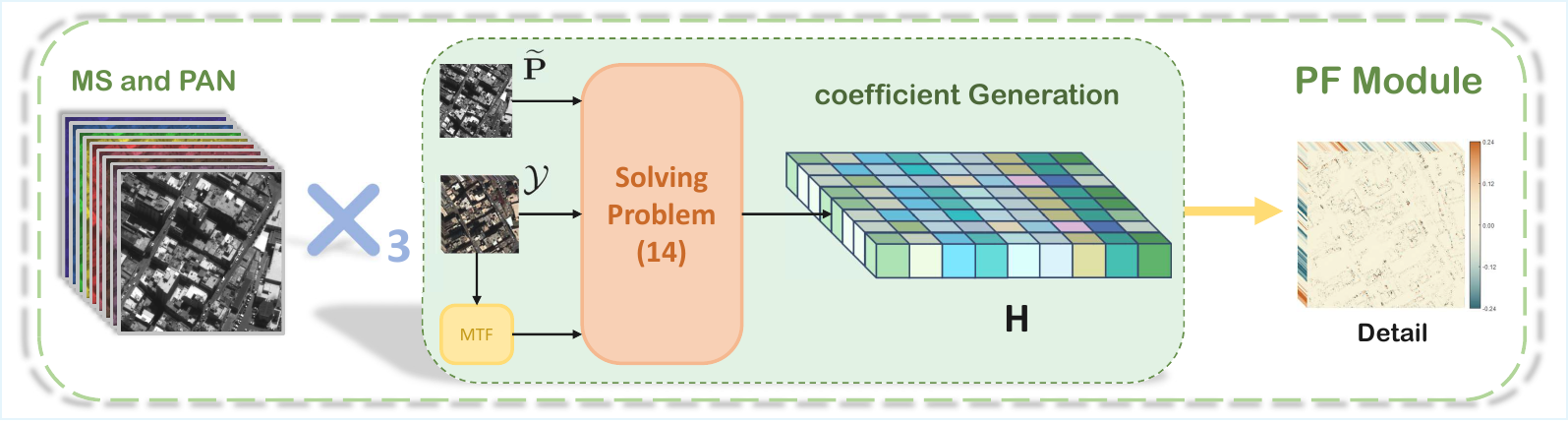}
  \caption{Flowchart of the PF module. At the downsampled resolution, the coefficient matrix $\mathbf{H}$ is obtained by solving Eq. \eqref{eq: reduced_4}. The reconstructed LRMS $\widehat{\mathcal{Y}}$ is then multiplied with $\mathbf{H}$ along mode-3 to generate the reference detail image for guiding the fusion process.}
  \label{PF_module}
\end{figure*}

\begin{figure}
    \centering
    \includegraphics[width=1\linewidth]{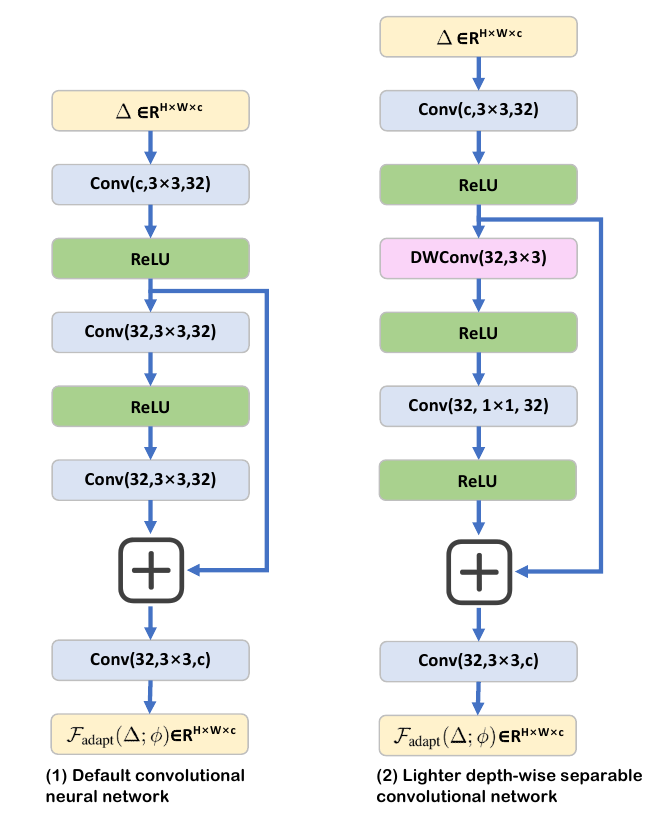}
    \caption{Comparison of two adaptive models. (1) Standard conv backbone with residual blocks. (2) Lightweight variant using depthwise separable convolutions and fewer channels for reduced complexity. Model (1) is used by default. DWConv$(\cdot,\cdot)$ is the depthwise separable convolution and Conv$(\cdot,\cdot,\cdot)$ is the standard convolution.}
    \label{Adaptive_model}
\end{figure}

\subsection{Adaptive Model}
The adaptive model serves as the core learnable component in our framework, responsible for image-specific adaptation under a self-supervised training scheme. Given the nature of our task-adapting from a single PAN-LRMS image pair without access to large-scale datasets, we adopt a relatively lightweight design that strikes a balance between expressiveness and efficiency.

As illustrated in Fig.~\ref{Adaptive_model}, the default adaptive model follows an encoder-residual-decoder architecture. It consists of an initial convolutional layer for shallow feature extraction, followed by a single residual block, and a final reconstruction layer to generate the HRMS output. This minimal backbone is sufficient to capture essential spatial-spectral cues from a single scene, enabling rapid convergence with limited computational cost.

The model input is constructed as the spectral difference between the upsampled LRMS image $\widehat{\mathcal{Y}}$ and a channel-wise replicated PAN image, denoted as $\mathcal{P}^D \in \mathbb{R}^{H \times W \times c}$. The difference tensor $\Delta$ is computed as:
\begin{equation}
\Delta = \mathcal{P}^D - \widehat{\mathcal{Y}},
\end{equation}
and serves as the input to the adaptive network, yielding the predicted HRMS image $\mathbf{X}^*$:
\begin{equation}
\mathcal{X}^* = \mathcal{F}_{\text{adapt}}(\Delta;\phi)+\widehat{\mathcal{Y}},
\end{equation}
where $\mathcal{F}_{\text{adapt}}(\cdot)$ denotes the adaptive model and $\phi$ its parameters.

In addition to the default model, Fig.~\ref{Adaptive_model} also presents a more lightweight variant that further reduces computational complexity while maintaining key adaptation capabilities. This variant employs depthwise separable convolutions and uses fewer parameters. While not used as the main architecture in this paper, its performance and trade-offs are analyzed in detail in \ref{ablation}.
Unless otherwise specified, all results and comparisons reported in this paper are based on the default adaptive model described above.

\subsection{Pretrained Model}
To leverage the powerful representation capability of deep networks, we employ a pretrained model as a reliable prior. This model is trained on reduced pansharpening datasets and possesses both good fidelity reconstruction performance and efficient inference speed. When given an input pair of PAN and LRMS images, it produces a high-quality HRMS image with strong spatial and spectral consistency.

In our framework, the pretrained model serves two purposes. First, it provides a high-quality pseudo-reference HRMS image $\mathcal{X}$ for guiding the adaptation process. Second, since the model is frozen during adaptation, it offers a stable supervisory signal that allows the adaptive model to converge rapidly, even though in a single-image training scenario.

To effectively utilize the pretrained output as supervision, we define the pseudo-reference loss, $\mathcal{L}_{\text{pr}}$, as the $\ell_2$ distance between the output of the adaptive model ($\mathcal{X}^{*}$) and that of the pretrained model ($\mathcal{X}$):
\begin{equation}
\mathcal{L}_{\text{pr}} = \| \mathcal{X}^{*} - \mathcal{X} \|_{F}^2,
\end{equation}
where $\| \cdot \|_{F}^2$ is the $\ell_2$ norm.

This loss encourages the adaptive model to mimic the structure and spectral information captured by the pretrained network, fostering stable and accurate local adaptation.

\subsection{Spectral Loss}
To ensure that the generated HRMS image maintains spectral consistency with the input LRMS image, we introduce a spectral fidelity loss. This loss measures the deviation between the LRMS input and a reduced-resolution version of the predicted HRMS image (i.e., the consistency property).

Specifically, the predicted HRMS image is first passed through a sensor degradation simulation, including a blurring step modeled by the multispectral sensor's MTF, followed by spatial decimation to match the scale of the LRMS input. The spectral fidelity loss, $\mathcal{L}_{\text{spe}}$, is then formulated as:
\begin{equation}\label{eq: spe}
\mathcal{L}_{\text{spe}} = \| \mathcal{Y} - \left( \mathtt{MTF}(\mathcal{X}^{*}) \right)\downarrow \|_{F}^2,
\end{equation}
where $\mathcal{X}^{*}$ is the output HRMS image from the adaptive model, $\mathcal{Y}$ denotes the original LRMS image, $\mathtt{MTF}(\cdot)$ represents the sensor-specific spatial blurring, and the $\downarrow$ symbol indicates the decimation operation. The $\ell_2$ norm penalizes spectral differences between the LRMS input and the spatial reduced-resolution version of the generated HRMS image. Hence, this loss term encourages the network to retain the original spectral characteristics during the pansharpening process.

\begin{table*}[t]
  \centering 
    \caption{Performance comparison on real-world data of WorldView-3 (WV3) and WorldView-2 (WV2) datasets. The reported values represent the average performance over 20 test images. (\colorbox{red!10}{Red}: best; \colorbox{blue!10}{Blue}: second best)}
  \label{tab:wv3_wv2_results}
  \setlength{\tabcolsep}{4pt}
  \renewcommand\arraystretch{1.2}
  \resizebox{0.95\linewidth}{!}{
    \begin{tabular}{c|ccc|ccc}
      \toprule
      \toprule
      \multirow{2.2}{*}{Method} & \multicolumn{3}{c|}{WV3 (Real Data): Avg$\pm$std} & \multicolumn{3}{c}{WV2 (Real Data): Avg$\pm$std} \\
      \cmidrule{2-4}\cmidrule{5-7}
       & HQNR$\uparrow$ & $D_\lambda\downarrow$ & $D_s\downarrow$ & HQNR$\uparrow$ & $D_\lambda\downarrow$ & $D_s\downarrow$ \\
      \midrule
          BT-H $^{\textcolor{gray}{2017}}$        & 0.8659$\pm$0.0568 & 0.0656$\pm$0.0262 & 0.0742$\pm$0.0383 & 0.8300$\pm$0.0430 & 0.0860$\pm$0.0301 & 0.0925$\pm$0.0208 \\
          C-BDSD $^{\textcolor{gray}{2014}}$     & 0.8562$\pm$0.0233 & 0.0874$\pm$0.0236 & 0.0618$\pm$0.0128 & 0.6956$\pm$0.0461 & 0.2253$\pm$0.0488 & 0.1019$\pm$0.0296 \\
          BDSD-PC$^{\textcolor{gray}{2019}}$      & 0.8673$\pm$0.0543 & 0.0634$\pm$0.0246 & 0.0749$\pm$0.0359 & 0.8286$\pm$0.0432 & 0.1413$\pm$0.0320 & ${0.0356\pm0.0213}$ \\
          MTF-GLP $^{\textcolor{gray}{2006}}$      & 0.9026$\pm$0.0444 & 0.0373$\pm$0.0124 & 0.0628$\pm$0.0359 & 0.8549$\pm$0.0475 & 0.0582$\pm$0.0221 & 0.0930$\pm$0.0320 \\
          MTF-GLP-FS $^{\textcolor{gray}{2018}}$ & 0.9127$\pm$0.0348 & 0.0357$\pm$0.0106 & 0.0537$\pm$0.0273 & 0.8658$\pm$0.0415 & 0.0563$\pm$0.0212 & 0.0830$\pm$0.0260 \\
          MF $^{\textcolor{gray}{2016}}$               & 0.9014$\pm$0.0358 & 0.0452$\pm$0.0121 & 0.0561$\pm$0.0268 & 0.8508$\pm$0.0538 & 0.0704$\pm$0.0308 & 0.0857$\pm$0.0297 \\
          PsDip $^{\textcolor{gray}{2024}}$         & 0.9215$\pm$0.0176 & 0.0191$\pm$0.0078 & 0.0607$\pm$0.0117 & 0.8980$\pm$0.0226 & ${0.0385\pm0.0239}$ & 0.0659$\pm$0.0158 \\
          ZS-Pan $^{\textcolor{gray}{2024}}$      & 0.9449$\pm$0.0208 & 0.0254$\pm$0.0071 & 0.0306$\pm$0.0153 & ${0.9112\pm0.0336}$ & 0.0476$\pm$0.0270 & 0.0435$\pm$0.0130 \\
          WFANet $^{\textcolor{gray}{2025}}$        & 0.9437$\pm$0.0130 & $\best{0.0172\pm0.0097}$ & 0.0390$\pm$0.0077 & 0.9280$\pm$0.0240 & 0.0584$\pm$0.0089 & 0.0430$\pm$0.0203 \\
          DICNN $^{\textcolor{gray}{2019}}$  & 0.9076$\pm$0.0288& 0.0487$\pm$0.0148& 0.0462$\pm$0.0171& 0.7137$\pm$0.0120 & 0.1369$\pm$0.1731 & 0.0290$\pm$0.0170 \\
          LAGNet $^{\textcolor{gray}{2022}}$         & 0.9133$\pm$0.0291 & ${0.0389\pm0.0133}$ & 0.0500$\pm$0.0196 & 0.8265$\pm$0.1356 & 0.0438$\pm$0.0458 & 0.1692$\pm$0.0580 \\
          FusionMamba $^{\textcolor{gray}{2024}}$ & $\second{0.9550\pm0.0108}$& $\second{0.0183\pm0.0076}$& ${0.0272\pm0.0057}$& 0.9064$\pm$0.0226& 0.0526$\pm$0.0252& 0.0432$\pm$0.0114\\
          \midrule
          FMG-Pan$_{\text{LAGNet}}$     & 0.9522$\pm$0.0195 & 0.0230$\pm$0.0065 &$\second{ 0.0255\pm0.0142}$ &$\second{0.9401\pm0.0193}$ & $\second{0.0332\pm0.0173}$ & $\second{0.0276\pm0.0490}$ \\
          FMG-Pan$_{\text{FusionMamba}}$        & $\best{0.9637\pm0.0116}$& 0.0213$\pm$0.0059& $\best{0.0153\pm0.0077}$& $\best{0.9471\pm0.0153}$ & \best{0.0315$\pm$0.0171} & $\best{0.0221\pm0.0048}$ \\
      \bottomrule
      \bottomrule
    \end{tabular}
  }
\end{table*}

\subsection{Physical Loss}
To further ensure the spatial consistency with the input PAN image and make more full use of the input information, we propose a novel physical loss.

The PAN image contains spatial details that are critical for the target HRMS image. However, the detail extraction is challenging since the relationship between HRMS and PAN images is unclear. To describe the above relationship, the band-dependent spatial-detail (BDSD) framework could provide a general scheme for detail preservation:
\begin{equation}
\begin{split}
\mathbf{X}_{i} = \widehat{\mathbf{Y}}_{i} + g_{i}\cdot\left(\mathbf{P}-\sum_{k=1}^{c} w_{k,i}\cdot \widehat{\mathbf{Y}}_{k}\right), \qquad i=1, \cdots, c,
\end{split}
\label{eq:BDSD}
\end{equation}
where $\widehat{\mathbf{Y}}_{i}$ denotes the $i$-th spectral band of the upsampled LRMS image $\widehat{\mathcal{Y}}$, and $\mathbf{X}_{i}$ is the $i$-th spectral band of the BDSD fused result. $w_{k,i}$ and $g_{i}$ ($k, i = 1, \cdots c$) are injection coefficients that match the high-frequency details of PAN and HRMS images. While this classic framework has been proven effective in~\citep{garzelli2007optimal,C-BDSD,BDSD-PC}, the upsampling operation inevitably introduces wrong information, which leads to inaccuracy of spatial information. However, this constraint cannot help optimize the adaptive model well, so we propose to just use its high-frequency detail information. To accurately describe the spatial relationship, we propose a novel physical fidelity as follows:
\begin{equation}
\begin{split}
\mathbf{X}_{i} = \mathbf{X}_{i}\mathbf{B} + g_{i}\cdot\left(\mathbf{P}-\sum_{k=1}^{c} w_{k,i}\cdot \widehat{\mathbf{Y}}_{k}\right), \qquad i=1, \cdots, c,
\end{split}
\label{eq:our_PF}
\end{equation}
where $\mathbf{B}\in \mathbb{R}^{HW \times HW}$ is the blurring operation. For convenience, we reformulate the fidelity as:
\begin{equation}
\begin{split}
\mathcal{X} - \mathtt{MTF}(\mathcal{X}) = \mathcal{M}\times_3 \mathbf{H}, 
\end{split}
\label{eq: new}
\end{equation}
where $\mathcal{M}\in \mathbb{R}^{H\times W\times (c+1)}$ denotes the concatenation of $\widehat{\mathbf{Y}}_{k}$ and $\mathbf{P}$, $\times_3$ is the mode-3 multiplication, and $\mathbf{H}\in \mathbb{R}^{c\times(c+1)}$ is the coefficient, which satisfies:
\begin{equation}
\begin{split}
\mathbf{H}(i,j) = \left\{
\begin{array}{lr}
-g_i w_{j,i}, \qquad & j=1,\cdots,c,\\
g_i, & j=c+1.
\end{array}
\right.
\end{split}
\end{equation}

Thus, we can build the physical loss, $\mathcal{L}_{\text{phy}}$, as:
\begin{equation}
\begin{split}
\mathcal{L}_{\text{phy}} = \| \mathcal{X}^{*} - \mathtt{MTF}(\mathcal{X}^{*}) - \mathcal{M}\times_3 \mathbf{H}\|_{F}^2.
\end{split}
\end{equation}

We denote the right side of the equation \eqref{eq: new} as $\mathbf{Detail}$ and the left side as $\mathbf{Detail^*}$.
The coefficient $\mathbf{H}$ is critical for connecting the detail of the PAN image. To estimate it, we built the corresponding relationship at reduced scale. Thus, we have:
\begin{equation}\label{eq: reduced_1}
\begin{split}
(\mathbf{X}_{(3)}\mathbf{B})\downarrow - ((\mathbf{X}_{(3)}\mathbf{B}) \mathbf{B})\downarrow  = (\mathbf{H}(\mathbf{M}_{(3)}\mathbf{B}))\downarrow,
\end{split}
\end{equation}
where $\mathbf{X}_{(3)}$ and $\mathbf{M}_{(3)}$ denote the mode-$3$ unfolding of $\mathcal{X}$ and $\mathcal{M}$, respectively. Since $\mathcal{X}$ is unavailable, we can assume there exist a Gaussian error $\epsilon$ satisfying the following equation~\citep{wu2023lrtcfpan}:
\begin{equation}
\begin{split}
((\mathbf{X}_{(3)}\mathbf{B})\mathbf{B})\downarrow  = (\mathbf{X}_{(3)}\mathbf{B}+\epsilon)\downarrow\mathbf{B}.
\end{split}
\end{equation}
Hence, equation \eqref{eq: reduced_1} can be rewritten as:
\begin{equation}\label{eq: reduced_2}
\begin{split}
(\mathbf{X}_{(3)}\mathbf{B})\downarrow - (\mathbf{X}_{(3)}\mathbf{B}+\epsilon)\downarrow\mathbf{B} = (\mathbf{H}(\mathbf{M}_{(3)}\mathbf{B}))\downarrow.
\end{split}
\end{equation}
Combining the spectral fidelity \eqref{eq: spe}, we have the reduced scale spatial relationship as follows:
\begin{equation}\label{eq: reduced_3}
\begin{split}
\mathbf{H}\mathbf{M}_\text{RD}=&(\mathbf{X}_{(3)}\mathbf{B})\downarrow - (\mathbf{X}_{(3)}\mathbf{B}+\epsilon)\downarrow\mathbf{B} \\\approx&\mathbf{Y}_{(3)}-\mathbf{Y}_{(3)}\mathbf{B},
\end{split}
\end{equation}
where $\mathbf{Y}_{(3)}$ denotes the mode-$3$ unfolding of $\mathcal{Y}$, and $\mathbf{M}_\text{RD}$ means $\mathbf{M}_{(3)}$ after the downsampling operation using MTF-based filters. Besides, following~\citep{BDSD-PC}, we estimate the coefficient with non-negative constraints. Finally, the coefficient $\mathbf{H}$ can be estimated by solving the problem as follows:
\begin{equation}\label{eq: reduced_4}
\begin{split}
&\min_{\mathbf{H}} \|\mathbf{H}\mathbf{M}_\text{RD}-(\mathbf{Y}_{(3)}-\mathbf{Y}_{(3)}\mathbf{B}) \|_{F}^2\\
&s.t. ~~g_i w_{j,i}~, g_i \geq 0, 
\end{split}
\end{equation}
which can be solved by the Newton method for the quadratic functions subject to bounds on some variables~\citep{coleman1996reflective}. The purpose of the PF module is to pre-generate Detail. Fig.~\ref{PF_module} shows its flowchart. It is worth noting that we can crop the original image pairs to estimate the coefficient faster, as we did in the experimental section.

Accordingly, the total training loss is defined as the weighted sum of the above three terms:
\begin{equation}
\mathcal{L}_{\text{total}} = \lambda_1 \mathcal{L}_{\text{pr}} + \lambda_2 \mathcal{L}_{\text{spe}} + \lambda_3 \mathcal{L}_{\text{phy}}.
\end{equation}

\section{Experiments}
\label{experiments}

\begin{figure*}[ht]
    \centering    
    \includegraphics[width=1\linewidth]{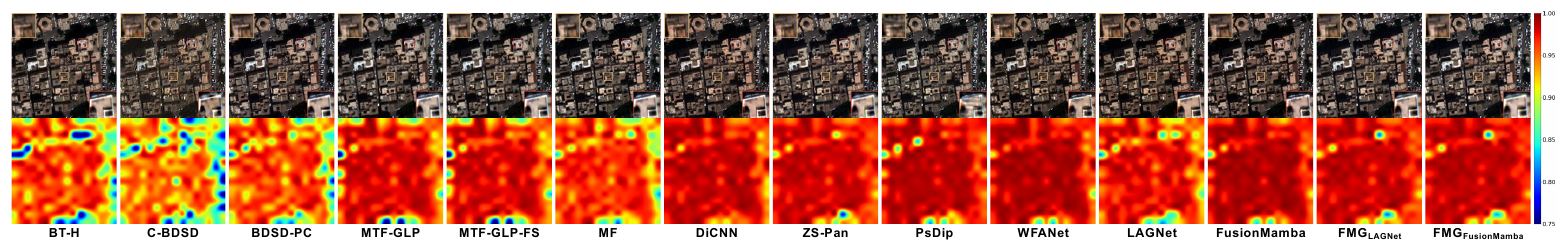}
    \caption{Visual fusion images and related HQNR maps on a full resolution WV3 example.}
    \label{fig:wv3_results}
\end{figure*}

\subsection{Datasets, Metrics, and Implementation Details}

\subsubsection{Datasets}
Due to our specific training strategy and loss function design, our task is targeted to real-world pansharpening, thus we adopted full resolution datasets in main experiments. In our experiments, we adopted two widely used satellite datasets: WorldView-3 (WV3), WorldView-2 (WV2). These datasets consist of paired PAN and LRMS images, with PAN images uniformly cropped to $512 \times 512$ pixels at real full resolution. At simulated reduced resolution, we have a size of $256 \times 256$ for PAN images. All datasets have a resolution ratio $r = 4$ between PAN and MS images.All datasets are derived from the PanCollection database. Details and related data can be found at\footnote{\url{https://github.com/liangjiandeng/PanCollection}}.


For the training of pretrained methods, we employ only the downsampled samples from WV3. WV2 are used solely as cross-sensor test sets to assess generalization. 
\begin{figure*}[htbp!]
    \centering
    \includegraphics[width=1\linewidth]{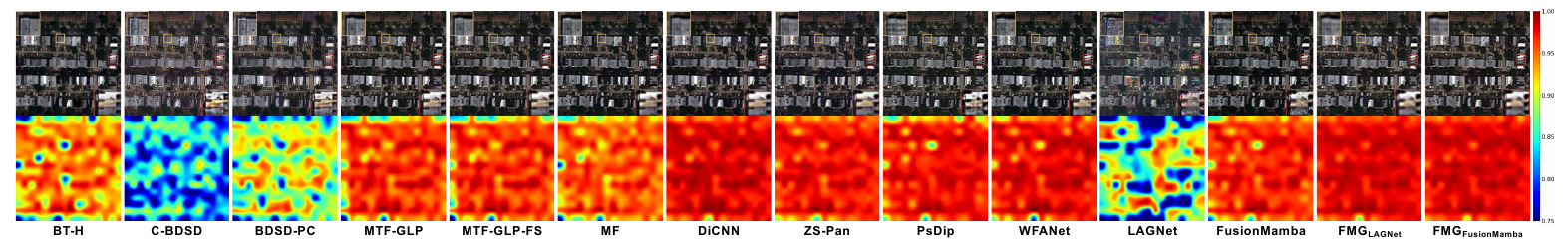}
    \caption{Visual fusion images and the related HQNR maps on a full resolution WV2 example.}  
    \label{fig:wv2_results} 
\end{figure*}
\subsubsection{Evaluation Metrics}
 To quantitatively assess pansharpening performance, we adopt widely recognized evaluation metrics. For real-world images, we utilize the HQNR index~\cite{aiazzi2014full}, which is derived from two complementary components: the spatial distortion metric $D_s$ and the spectral distortion metric $D_\lambda$. These three metrics jointly reflect image quality in the absence of ground-truth. Ideal values are $D_s=0$, $D_\lambda=0$, and $\text{HQNR}=1$.

\subsubsection{Implementation Details}
We use two pretrained networks as pretrained models in our framework: FusionMamba~\citep{peng2024fusionmamba} and LAGNet~\citep{jin2022lagconv}, which are pretrained on WV3. To ensure fairness and validate generalization performance, all trainable networks in our framework are trained only on the training splits of WV3 and GF2, while evaluations are conducted on both in-domain WV3 and cross-sensor WV2 test sets.

The adaptive model is initialized with pytorch default weights and optimized using the Adam optimizer with a learning rate of $1.8 \times 10^{-3}$. The number of epochs for training on WV3 is set to 80, and the number of epochs for the cross-sensor dataset WV2 is 350. The loss weights are set as follows: $\lambda_1 = 1$, $\lambda_2 = 0.5$, and $\lambda_3 = 10$. To accelerate inference in the PF module, we compute the coefficient using only a small central patch (64$\times$64 for the PAN image size) rather than the entire image. This significantly improves runtime efficiency.

For baseline comparisons, we adopt the default configurations specified in the original papers and codebases of each deep learning-based method. All the experiments were conducted on a hardware setup comprising an NVIDIA RTX 3090 GPU with 24GB memory and Intel Xeon Gold 6148 CPU.

\subsection{Benchmark}
We compare FMG-Pan with some representative pansharpening techniques, including

\textbf{Traditional methods:} MTF-GLP: MTF-tailored multiscale fusion of high-resolution MS and Pan imagery~\citep{aiazzi2006mtf}. C-BDSD: Pansharpening of multispectral images based on nonlocal parameter optimization~\citep{C-BDSD}. MF: Fusion of multispectral and panchromatic images based on morphological operators~\citep{MF}. BT-H: Haze correction for contrast-based multispectral pansharpening~\citep{BT-H}. MTF-GLP-FS: Full scale regression-based injection coefficients for panchromatic sharpening~\citep{MTF-GLP-FS}. BDSD-PC: Robust band-dependent spatial-detail approaches for panchromatic sharpening~\citep{BDSD-PC}.

\textbf{Supervised DL methods:} DICNN: Pansharpening via detail injection based convolutional neural networks~\citep{DICNN}. LAGNet: Local-context adaptive convolution kernels with global harmonic bias for pansharpening~\citep{jin2022lagconv}. FusionMamba: Efficient remote sensing image fusion with state space model~\citep{peng2024fusionmamba}. WFANet: Wavelet-assisted multi-frequency attention network for pansharpening~\citep{WFANet}.

\textbf{Zero-shot methods:} PsDip: Variational zero-shot multispectral pansharpening~\citep{PsDip}. ZS-Pan: Zero-shot semi-supervised learning for pansharpening~\citep{ZsPan}.

\subsection{Main Results}
To evaluate the effectiveness of our method, we report the main real-world results on WV3 and WV2 and compare against representative pansharpening techniques. models are trained on WV3 and the quantitative results are reported in Table~\ref{tab:wv3_wv2_results}.

The HQNR, $D_\lambda$, and $D_s$ metrics show that even when the pretrained model itself is not the strongest performer, the FMG-Pan framework significantly enhances its performance. In particular, when using FusionMamba as the pretrained model, our framework achieves the best overall metrics on both WV3 and WV2, highlighting strong real-world performance and cross-sensor generalization.

Figs.~\ref{fig:wv3_results} and~\ref{fig:wv2_results} further illustrate the fused results and HQNR heatmaps (with local zoom-in regions). The visual comparisons confirm that FMG-Pan delivers higher-quality fusion, simultaneously preserving fine spatial details and spectral fidelity.

\begin{figure}
    \centering
    \includegraphics[width=1\linewidth]{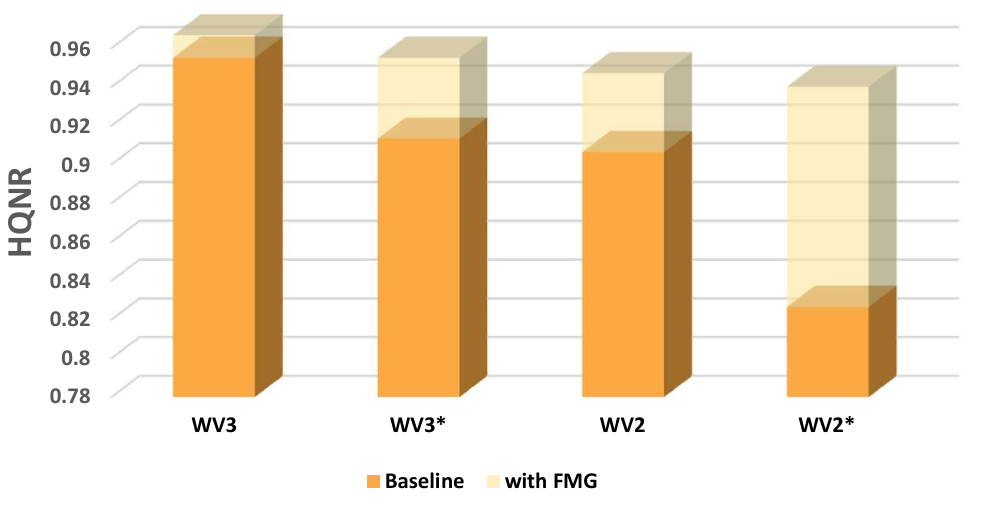}
    \caption{Comparison of HQNR for FusionMamba and LAGNet with and without the FMG-Pan framework across two real-world datasets (WV3, WV2). Note: * denotes LAGNet, without * denotes FusionMamba.}
    \label{fig:HQNR_CMP}
\end{figure}

In addition, Fig.~\ref{fig:HQNR_CMP} illustrates the improvement of the FMG-Pan framework when applied to FusionMamba and LAGNet across two datasets. The consistent gains in the overall HQNR metric clearly highlight both the effectiveness of our design and its strong generalization ability under diverse sensor conditions.

\subsection{Ablation Study}
\label{ablation}
We evaluated the contribution of each component in our framework on 20 full resolution WV3 samples using FusionMamba as the pretrained model. Four settings were considered: full FMG-Pan,FMG-Pan with Lighter adaptive model, no pretrained model, no spectral loss, and no physical loss. Table~\ref{tab:Ablation} reports the mean and standard deviation of HQNR, $D_\lambda$, and $D_s$. The Full FMG-Pan and FMG-Pan$^L$ almost achieved all the best metrics.
\subsubsection{FMG-Pan with lighter adaptive model}
As shown in the table, even when employing a significantly lighter model—where the parameter count is reduced from 23k to 6k, performance remains nearly comparable to that of the full framework. This can be attributed to the fact that our framework is optimized solely for single-image processing.
Furthermore, this result highlights the practical potential of our approach: the adaptive model parameters involved in actual training can be further reduced without substantially compromising performance.

\subsubsection{Impact of Pretrained Model}

Removing the pretrained model resulted in a consistent performance drop in all metrics, see Table~\ref{tab:Ablation}. HQNR decreased from $0.9637$ to $0.9491$, while $D_\lambda$ and $D_s$ both increased. This confirms that model-guided initialization provides a strong prior for our single-image adaptation: it stabilizes the optimization process, accelerates convergence, and enables the adaptive model to overcome the original output of the pretrained model.

\subsubsection{Impact of Spectral Fidelity Loss}
Discarding the spectral loss significantly compromises the balance between spatial and spectral quality. HQNR drops to 0.9308, while $D_\lambda$ increases and $D_s$ swells from 0.0153 to 0.0473 (over 3 times), leading to oversharpening when spectral consistency is not explicitly constrained. Therefore, the spectral term is crucial for maintaining cross-band consistency.

\subsubsection{Impact of Physical Fidelity Loss}
Without the physical fidelity loss, $D_\lambda$ decreases slightly (to $0.0209$), but $D_s$ increases (from $0.0153$ to $0.0193$), and HQNR drops to $0.9602$. This trade-off demonstrates the importance of physical constraints in further balancing spatial and spectral fidelity.

\begin{table}[!ht]
\centering
\setlength{\tabcolsep}{1pt}
\renewcommand\arraystretch{1}
\caption{Average results of the ablation study for our FMG-Pan framework on 20 full-resolution WV3 examples. 
(\colorbox{red!10}{Red}: best; \colorbox{blue!10}{Blue}: second best)}
\label{tab:Ablation}

\resizebox{\linewidth}{!}{\begin{tabular}{l|ccc}
\toprule
\textbf{Name} & \textbf{HQNR}$\uparrow$ & $\boldsymbol{D_\lambda}\downarrow$ & $\boldsymbol{D_s}\downarrow$ \\
\midrule
$\text{FMG-Pan}$ & 
\colorbox{red!10}{{0.9637$\pm$0.0116}} &
\colorbox{blue!10}{{0.0213$\pm$0.0059}} &
\colorbox{blue!10}{{0.0153$\pm$0.0077}} \\
$\text{FMG-Pan}^\text{L}$ & \colorbox{blue!10}{{0.9636$\pm$0.0150}} & 0.0227$\pm$0.0056 & \colorbox{red!10}{{0.0140$\pm$0.0107}}\\
w/o pretrained model & 
0.9491$\pm$0.0128 & 0.0227$\pm$0.0059 & 0.0288$\pm$0.0085 \\

w/o spectral loss &
0.9308$\pm$0.0298 & 0.0232$\pm$0.0097 & 0.0473$\pm$0.0217 \\

w/o physical loss &
{{0.9602$\pm$0.0078}} &
\colorbox{red!10}{{0.0209$\pm$0.0058}} &
{{0.0193$\pm$0.0039}} \\
\bottomrule
\end{tabular}}
\end{table}

\subsection{Running Time and Efficiency}
We analyze the running time of our framework. As summarized in Table~\ref{tab:time}, FMG-Pan achieves superior pansharpening quality while being significantly faster than existing zero-shot methods. On WV3, FMG-Pan$_{\text{FusionMamba}}$ reaches an HQNR of 0.9637 with $D_\lambda=0.0213$ and $D_s=0.0153$ in only 2.93\,s per full resolution image, whereas ZS-Pan obtains an HQNR of 0.9449 in 67.24\,s and PsDip reaches an HQNR of 0.9215 in 285.37\,s. On WV2, our framework similarly delivers an HQNR of 0.9471 ($D_\lambda=0.0315$, $D_s=0.0221$) within 9.75\,s, outperforming ZS-Pan (HQNR of 0.9112, 68.52\,s) and PsDip (HQNR of 0.8980, 280.23\,s). When LAGNet is used as the pretrained model, the runtime remains comparable (2.84\,s on WV3; 9.62\,s on WV2) with only a slight decrease in accuracy relative to FusionMamba, suggesting that the efficiency stems from the FMG-Pan framework itself rather than the choice of the backbone.

We do not report comparisons with conventional supervised deep learning pipelines, as their training on datasets usually takes hours to days, making them not directly comparable to our single-image adaptation scheme, although they often have fast inference speed in the test time without adaptation.

\begin{table}[ht]
    \centering
        \caption{Average performance comparison of zero-shot methods and the proposed FMG-Pan framework on 20 full resolution test images for both WV3 and WV2 datasets. (\colorbox{red!10}{Red}: best; \colorbox{blue!10}{Blue}: second best)}
    \setlength{\tabcolsep}{6pt}
    \renewcommand\arraystretch{1.2}
    \resizebox{\linewidth}{!}{
    \begin{tabular}{c|l|cccc}
         \toprule
         Dataset & Method & HQNR$\uparrow$ & \(D_\lambda\downarrow\) & \(D_s\downarrow\) & Time$\downarrow$ (s) \\
         \midrule
         \multirow{4}{*}{WV3} 
           
         & PsDip             & 0.9215 & \best{0.0191} & 0.0607 & 285.37    \\
         & ZS-Pan            & 0.9449 & 0.0254 & 0.0306 & 67.24    \\
         & $\text{FMG-Pan}_\text{LAGNet}$   & \second{0.9522} & 0.0230 &$\second{ 0.0255}$ & \best{2.84}  \\
         & $\text{FMG-Pan}_\text{FusionMamba}$       & \best{0.9637} & \second{0.0213} & \best{0.0153} & \second{2.93} \\

         \midrule
         \multirow{4}{*}{WV2}
         & PsDip           &0.8980 & {0.0385} & 0.0659 & 280.23    \\
         & ZS-Pan            &{0.9112} & 0.0476 & 0.0435 & 68.52    \\
         & $\text{FMG-Pan}_\text{LAGNet}$   &$\second{0.9401}$ & $\second{0.0332}$ & $\second{0.0276}$ & \best{9.62}  \\
         & $\text{FMG-Pan}_\text{FusionMamba}$       &\best{0.9471} & \best{0.0315} & \best{0.0221} & \second{9.75} \\

         \bottomrule
    \end{tabular}
    }
    \label{tab:time}
\end{table}

\section{Conclusion}
\label{conclusion}
In this paper, we proposed a model-guided Instance adaptation framework to address two major challenges in pansharpening: the poor generalization ability of conventional deep learning methods and the low inference efficiency of existing single-image training approaches. Our framework introduced a novel physical fidelity term, which, together with the spectral fidelity term, guides the adaptive model training under the supervision of a pretrained network. This design enables the framework to achieve high performance and strong generalization on real-world datasets while maintaining fast inference speed.
Extensive experiments on multiple satellite datasets demonstrated that the proposed FMG-Pan gets state-of-the-art results with rapid convergence, improving both the model's accuracy and robustness on real-world datasets, highlighting its potential for real-world pansharpening. 


\bibliographystyle{ACM-Reference-Format}
\bibliography{main}

\clearpage
\appendix
\renewcommand{\thesection}{S\arabic{section}}
\setcounter{section}{0}
\twocolumn[{%
\begin{center}
  {\fontsize{20}{24}\selectfont\bfseries Supplementary Material\par}
  \vspace{1.5em}
\end{center}
}]
\section{Time Analysis}

\begin{figure}[h]
    \centering
    \includegraphics[width=1\linewidth]{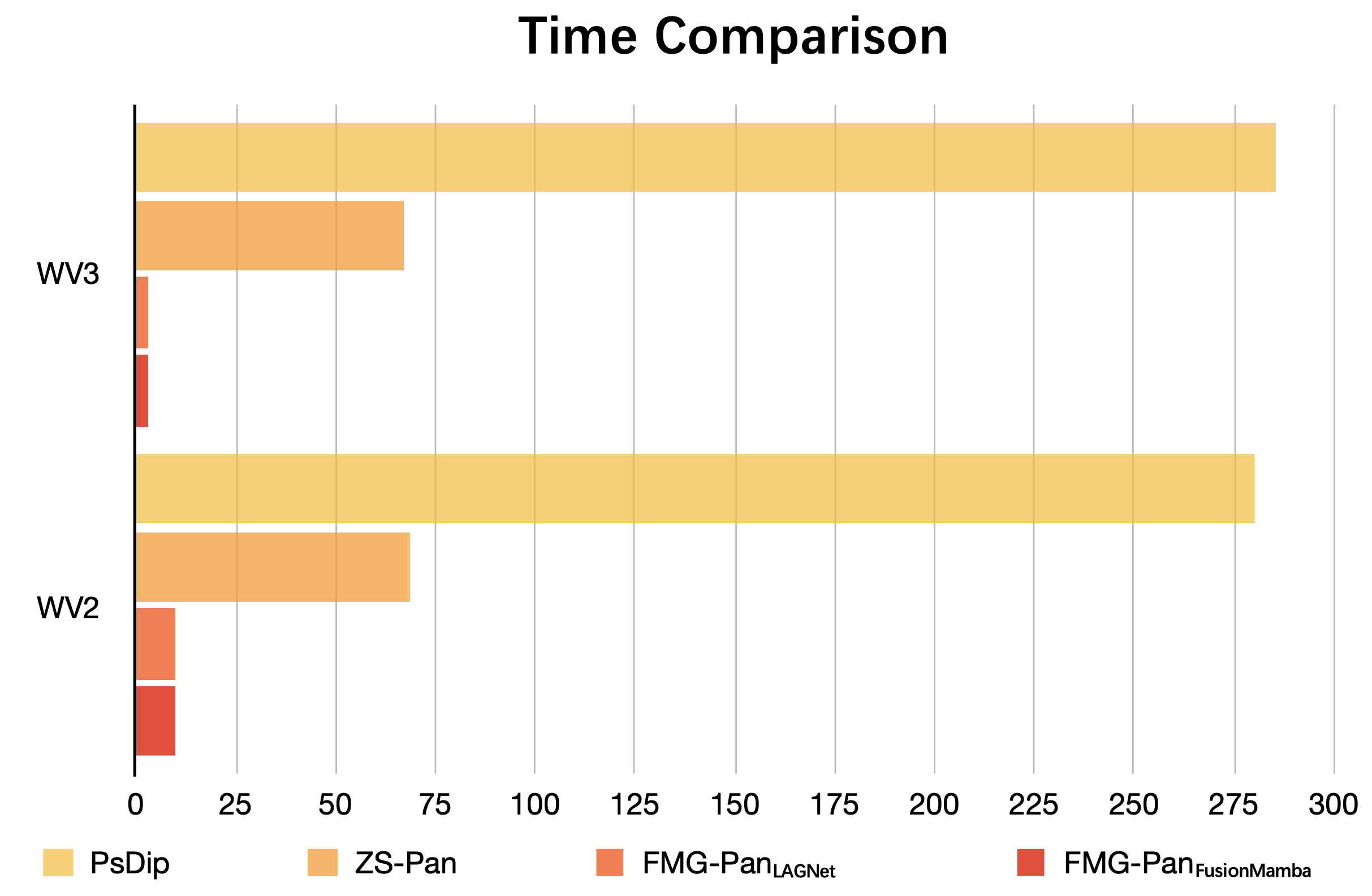}
    \caption{Time comparison (in seconds) between the proposed FMG-Pan framework (with LAGNet and FusionMamba as pretrained models) and previous zero-shot methods (PsDip, ZS-Pan) on WV3 and WV2 data.}
    \label{fig:Time Comparison}
\end{figure}

\begin{figure}
    \centering
    \includegraphics[width=1\linewidth]{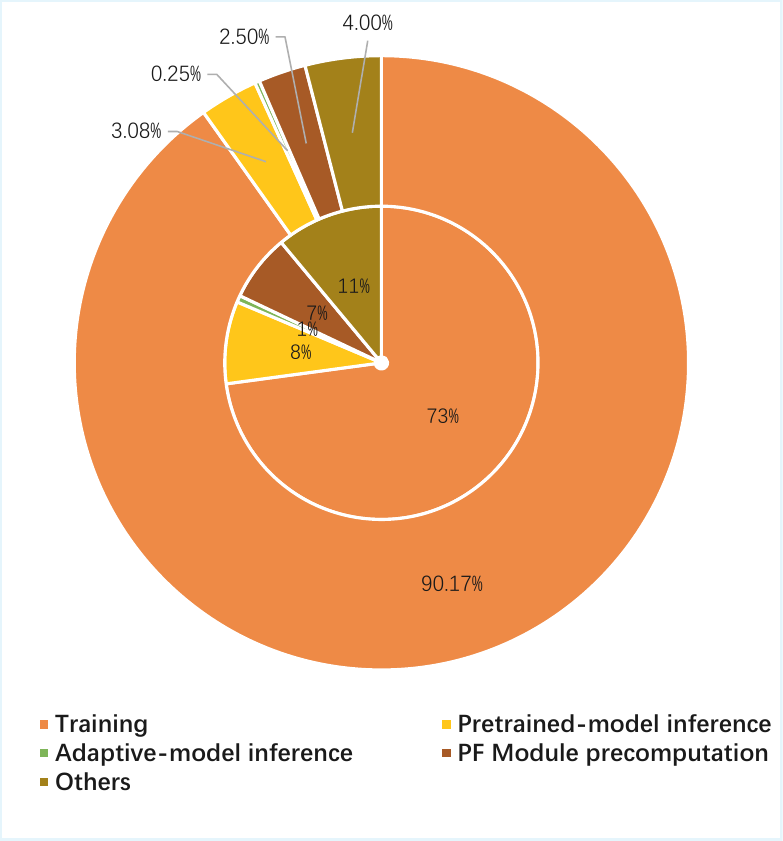}
    \caption{Runtime composition of the proposed FMG-Pan framework using FusionMamba as the pretrained model. The inner ring corresponds to WV3 and the outer ring to WV2, showing the proportion of time spent on training, pretrained model inference, adaptive-model inference, PF module precomputation, and other processes.}
    \label{fig:Time}
\end{figure}

Fig.~\ref{fig:Time Comparison} shows our efficiency advantage compared to previous zero-shot methods.
Fig.~\ref{fig:Time} provides a breakdown of runtime composition when using FusionMamba as the pretrained model. For WV3 (inner ring), training accounts for 73\% of the total time, while pretrained model inference, PF module precomputation, adaptive-model inference, and other processes account for 8\%, 7\%, 1\%, and 11\%, respectively. For WV2 (outer ring), training becomes even more dominant at 90.17\%, with pretrained inference at 3.08\%, PF module precomputation 2.50\%, adaptive-model inference 0.25\%, and others 4.00\%. These results confirm that the majority of time is spent on the single-image adaptation, whereas both inference and precomputation are relatively minor. The total time is consistent with the overall runtime, where FMG-Pan completes the full resolution inference within only a few seconds.
\section{Additional Main Results on GF2 and QB}
To evaluate the effectiveness of our method, we conduct experiments on the 4-band in-domain dataset GF2 and the cross-domain dataset QB using the same configuration. The quantitative results are reported in Table~\ref{tab:gf2_qb_results}, where our method is compared with several representative pansharpening approaches.

The evaluation metrics demonstrate that our model consistently achieves superior performance across both datasets. In particular, our method yields competitive or leading results in terms of key quantitative indicators, indicating its strong capability in both in-domain reconstruction and cross-domain generalization.

\begin{figure}[!ht]
    \centering
    \includegraphics[width=1\linewidth]{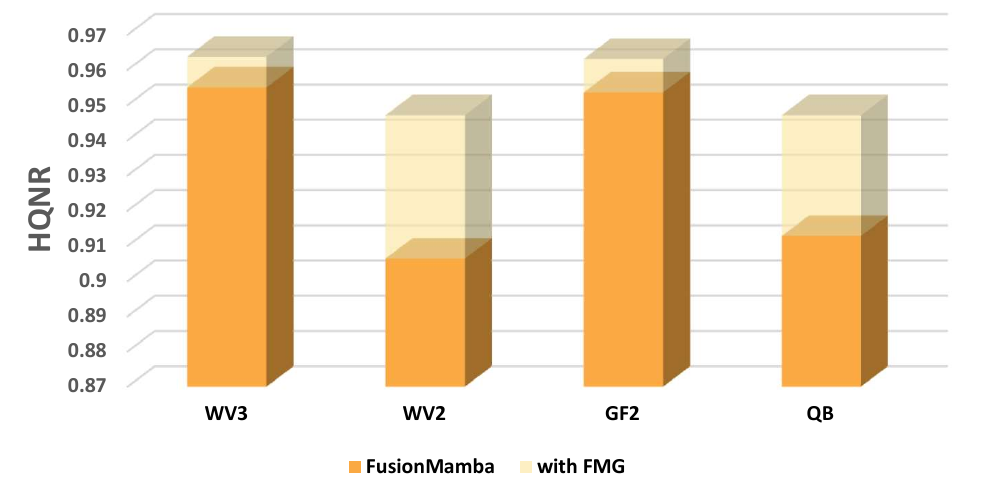}
    \caption{Comparison of HQNR for FusionMamba with and without the FMG-Pan framework across four real-world datasets (WV3, WV2, GF2, QB)}
    \Description{}
    \label{fig:CMP_HQNR}
\end{figure}

Fig.~\ref{fig:CMP_HQNR} further compares the HQNR improvements across four datasets when using FusionMamba as the pretrained model. It can be observed that our framework consistently boosts the HQNR scores on all datasets, demonstrating its effectiveness in enhancing different backbone models.

\begin{figure}[!t]
    \centering
    \includegraphics[width=1\linewidth]{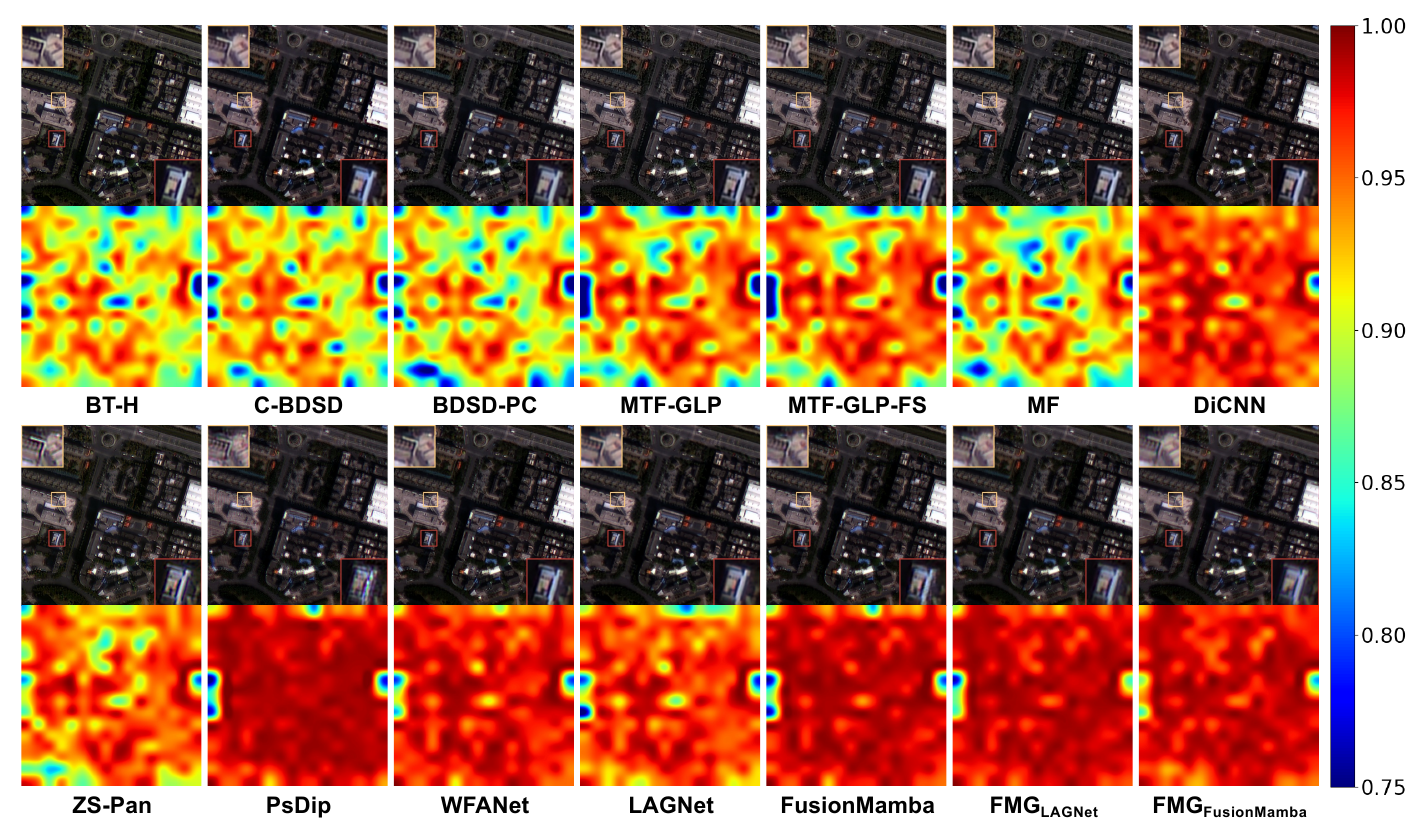}
    \caption{Visual fusion images and the related HQNR maps on a full resolution GF2 example.}
    \Description{A full-resolution GF2 qualitative comparison with HQNR maps.}
    \label{fig:gf2_results}
\end{figure}

\begin{figure}[!t]
    \centering
    \includegraphics[width=1\linewidth]{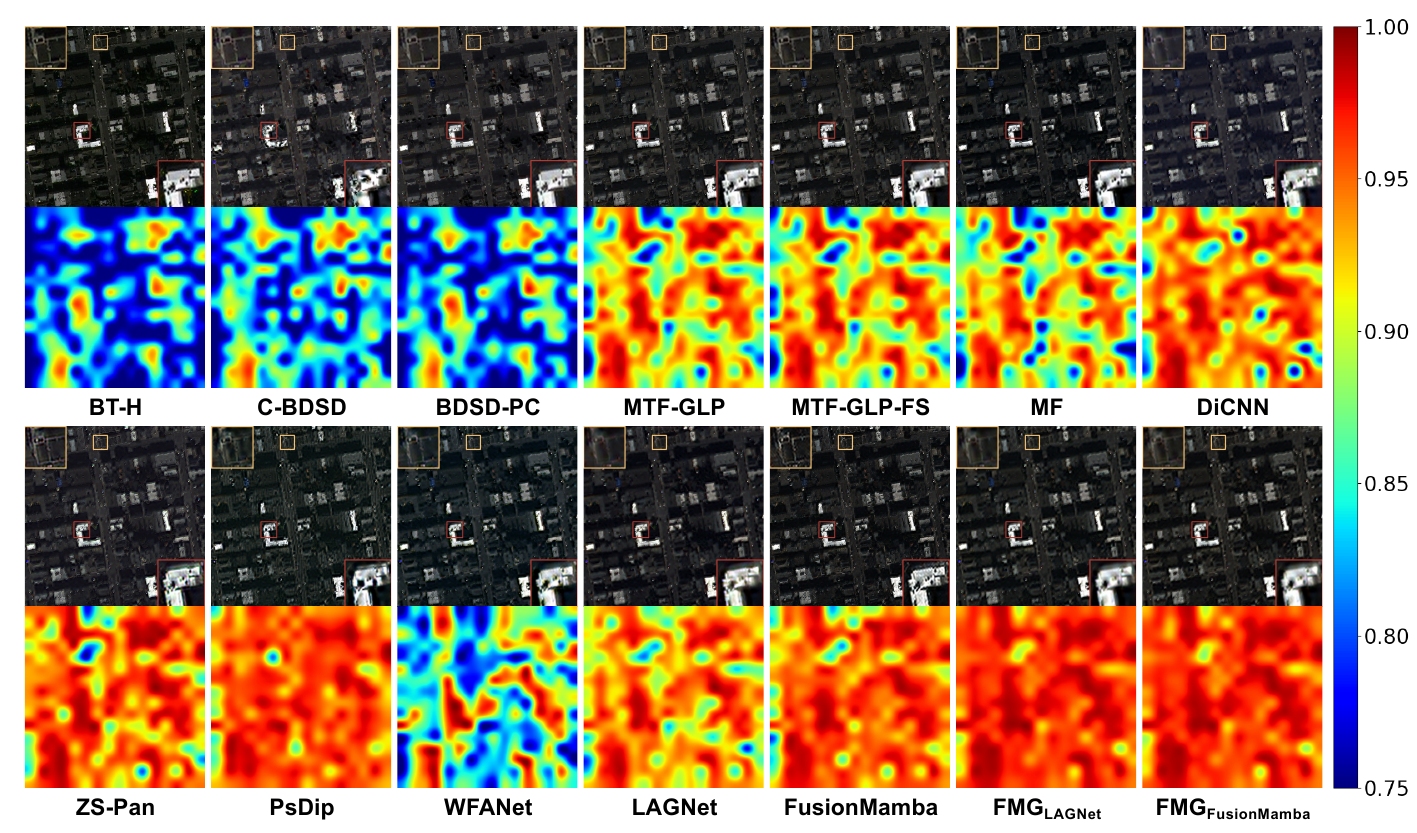}
    \caption{Visual fusion images and the related HQNR maps on a full resolution QB example.}
    \Description{A full-resolution QB qualitative comparison with HQNR maps.}
    \label{fig:qb_results}
\end{figure}

Figs.~\ref{fig:gf2_results} and~\ref{fig:qb_results} present the visual comparisons of the fused images. As shown in the zoomed-in regions, our method is able to better preserve fine spatial details while maintaining high spectral fidelity. Compared with other methods, it effectively reduces spectral distortion and spatial artifacts, resulting in more visually pleasing and structurally consistent fusion results.

Overall, our model achieves state-of-the-art performance on both GF2 and QB datasets in terms of quantitative metrics and visual quality, further validating its robustness and effectiveness in real-world scenarios.

\begin{table*}[!t]
  \centering 
    \caption{Performance comparison on real-world GaoFen-2 (GF2) and QuickBird (QB) data. The reported values represent the average performance over 20 test images. (\colorbox{red!10}{Red}: best; \colorbox{blue!10}{Blue}: second best)}
  \label{tab:gf2_qb_results}
  \setlength{\tabcolsep}{4pt}
  \renewcommand\arraystretch{1.2}
  \resizebox{\linewidth}{!}{
    \begin{tabular}{c|ccc|ccc}
      \toprule
      \toprule
      \multirow{2.2}{*}{Method} & \multicolumn{3}{c|}{GF2 (Real-World Data): Avg$\pm$std} & \multicolumn{3}{c}{QB (Real-World Data): Avg$\pm$std} \\
      \cmidrule{2-4}\cmidrule{5-7}
       & HQNR$\uparrow$ & $D_\lambda\downarrow$ & $D_s\downarrow$ & HQNR$\uparrow$ & $D_\lambda\downarrow$ & $D_s\downarrow$ \\
      \midrule
          BT-H $^{\textcolor{gray}{2017}}$        & 0.8923$\pm$0.0312 & 0.0832$\pm$0.0180 & 0.0814$\pm$0.0110 & 0.8901$\pm$0.0980 & 0.0457$\pm$0.0136 & 0.0693$\pm$0.0109 \\
          C-BDSD $^{\textcolor{gray}{2014}}$     & 0.8870$\pm$0.0210 & 0.0119$\pm$0.0094 & 0.0689$\pm$0.0079 & 0.8317$\pm$0.0750 & 0.0776$\pm$0.0385 & 0.0689$\pm$0.0089 \\
          BDSD-PC$^{\textcolor{gray}{2019}}$      & 0.8710$\pm$0.0087 & 0.0633$\pm$0.0078 & 0.0742$\pm$0.0095 & 0.8699$\pm$0.0670 & 0.0671$\pm$0.0074 & 0.0780${\pm}$0.0082 \\
          MTF-GLP $^{\textcolor{gray}{2006}}$      & 0.8247$\pm$0.0462 & 0.0598$\pm$0.0072 & 0.0545$\pm$0.0011 & 0.8574$\pm$0.0840 & 0.0311$\pm$0.0079 & 0.0546$\pm$0.0087 \\
          MTF-GLP-FS $^{\textcolor{gray}{2018}}$ & 0.8179$\pm$0.0912 & 0.0371$\pm$0.0015 & 0.0149$\pm$0.0099 & 0.8230$\pm$0.0970 & 0.0495$\pm$0.0058 & 0.1411$\pm$0.0083 \\
          MF $^{\textcolor{gray}{2016}}$               & 0.8730$\pm$0.1200 & 0.0646$\pm$0.0088 & 0.0633$\pm$0.0136 & 0.8429$\pm$0.0997 & 0.0453$\pm$0.0120 & 0.0686$\pm$0.0083 \\
          PsDip $^{\textcolor{gray}{2024}}$         & 0.8833$\pm$0.0221 & 0.0448$\pm$0.0212 & 0.0752$\pm$0.0205 & 0.7825$\pm$0.0325 & ${0.1323\pm0.0286}$ & 0.0981$\pm$0.0256 \\
          ZS-Pan $^{\textcolor{gray}{2024}}$      & 0.8961$\pm$0.0274 & 0.0361$\pm$0.0157 & 0.0705$\pm$0.0185 & ${0.8925\pm0.0240}$ & 0.0778$\pm$0.0182 & 0.0323$\pm$0.0159 \\
          WFANet $^{\textcolor{gray}{2025}}$        & 0.9463$\pm$0.0095 & $\best{0.0162\pm0.0078}$ & 0.0381$\pm$0.0042 & 0.8922$\pm$0.0223 & 0.0633$\pm$0.0221 & 0.0476$\pm$0.0073 \\
          DICNN $^{\textcolor{gray}{2019}}$  & 0.9076$\pm$0.0288& 0.0487$\pm$0.0148& 0.0462$\pm$0.0171& 0.8881$\pm$0.0213 & 0.0543$\pm$0.0273 & 0.0606$\pm$0.0162 \\
          LAGNet $^{\textcolor{gray}{2022}}$         & 0.8815$\pm$0.0293 & ${0.0423\pm0.0310}$ & 0.0795$\pm$0.0107 & 0.8879$\pm$0.0386 & 0.0814$\pm$0.0315 & 0.0336$\pm$0.0128 \\
          FusionMamba $^{\textcolor{gray}{2024}}$ & $\second{0.9536\pm0.0086}$& $\second{0.0174\pm0.0094}$ & $\second{0.0295\pm0.0073}$& 0.9128$\pm$0.0277& 0.0553$\pm$0.0208& 0.0339$\pm$0.0107\\
          \midrule
          FMG-Pan$_{\text{LAGNet}}$     & 0.9372$\pm$0.0156 & 0.0263$\pm$0.0136 & 0.0375$\pm$0.0127 & $\second{0.9447\pm0.0157}$ & $\best{0.0333\pm0.0117}$ & $\second{0.0227\pm0.0085}$ \\
          FMG-Pan$_{\text{FusionMamba}}$        & $\best{0.9631\pm0.0159}$ & 0.0228$\pm$0.0143 & $\best{0.0144\pm0.0057}$ & $\best{0.9471\pm0.0245}$ & $\second{0.0375\pm0.0165}$ & $\best{0.0161\pm0.0103}$ \\
      \bottomrule
      \bottomrule
    \end{tabular}
  }
\end{table*}

\begin{table*}[!t]
    \centering
    \caption{Comparison in terms of HQNR, $D_\lambda$, $D_s$, and runtime (s), of FMG-Pan$_{\text{FusionMamba}}$, its lighter-weight variant (FMG-Pan$^{\text{L}}_{\text{FusionMamba}}$), and the fine-tuned variant (FMG-Pan$^{\text{F}}_{\text{FusionMamba}}$) on 20 full resolution test images for both WV3 and WV2 datasets. (\colorbox{red!10}{Red}: best; \colorbox{blue!10}{Blue}: second best)}
    \setlength{\tabcolsep}{6pt}
    \renewcommand\arraystretch{1.2}
    \resizebox{\linewidth}{!}{
    \begin{tabular}{c|l|cccc}
         \toprule
         Dataset & Method & HQNR$\uparrow$ & \(D_\lambda\downarrow\) & \(D_s\downarrow\) & Runtime$\downarrow$ (s) \\
         \midrule
         \multirow{3}{*}{WV3}
         & $\text{FMG-Pan}_\text{FusionMamba}$       & \second{0.9637$\pm$0.0116} & 0.0213$\pm$0.0059 & \second{0.0153}$\pm$0.0077 & {2.93}$\pm$0.12 \\
         & $\text{FMG-Pan}_\text{FusionMamba}^\text{L}$ & {0.9636}$\pm$0.0150 & 0.0227$\pm$0.0056 & \best{0.0140}$\pm$0.0107 & \second{2.82}$\pm$0.13 \\
         & $\text{FMG-Pan}_\text{FusionMamba}^\text{F}$ & \best{0.9644}$\pm$0.0140 & \second{0.0205}$\pm$0.0052 & {0.0154}$\pm$0.0105 & \best{1.47}$\pm$0.08 \\
         \midrule
         \multirow{3}{*}{WV2}
         & $\text{FMG-Pan}_\text{FusionMamba}$       & \best{0.9471}$\pm$0.0193 & \best{0.0315}$\pm$0.0173 & \best{0.0221}$\pm$0.0490 & {9.75}$\pm$0.32 \\
         & $\text{FMG-Pan}_\text{FusionMamba}^\text{L}$ & \second{0.9460}$\pm$0.0256 & \second{0.0325}$\pm$0.0179 & \second{0.0223}$\pm$0.0096 & \second{9.55}$\pm$0.29 \\
         & $\text{FMG-Pan}_\text{FusionMamba}^\text{F}$ & {0.9439}$\pm$0.0252 & 0.0330$\pm$0.0182 & {0.0239}$\pm$0.0084 & \best{3.43}$\pm$0.16 \\
         \bottomrule
    \end{tabular}
    }
    \label{tab:wv3_wv2_results_runtime}
\end{table*}

\section{Lighter-weight Adaptive Model}
To additionally investigate the efficiency of our framework, we evaluate a lighter-weight variant of the adaptive model. We use FusionMamba as the pretrained model. The parameter count is reduced from 23k to 6k in the 8-band setting and from 20k to 3.7k in the 4-band setting. This substantial reduction in model size leads to lower computational cost and faster inference within the FMG-Pan framework.

As shown in Table~\ref{tab:wv3_wv2_results_runtime}, despite its extremely compact architecture, the lighter-weight adaptive model achieves fusion performance comparable to that of the original framework across two datasets while slightly improving computational speed. The HQNR, $D_\lambda$, and $D_s$ metrics indicate only marginal differences, suggesting that in our single-image adaptation setting, the adaptive model does not require a large number of parameters to effectively capture the spatial-spectral details of the PAN and LRMS inputs.

These results demonstrate that the proposed framework can operate efficiently with a highly compact adaptive module, making it a promising option for scenarios with limited computational resources or graphics memory.

\section{FMG-Pan with Fine-tuning}
In this experiment, we further investigate the potential to optimize the inference speed of our framework. For this experiment, we also use FusionMamba as the pretrained model. Inspired by the idea of fine-tuning, we can avoid training the adaptive model from scratch when we have the pretrained weights. For simplicity, we initialize the adaptive model on each dataset using the weights obtained from the single-image adaptation of the original FMG-Pan framework on the first sample (label 0) of that dataset, without performing full training on the corresponding training set.

We then fine-tune the adaptive model for only 30 epochs on the WV3 dataset, and for 100 epochs on the cross-sensor dataset (i.e., WV2). This setting could further reduce running times by reducing the epochs for the adaptive model on the training part.

As reported in Table~\ref{tab:wv3_wv2_results_runtime}, the fine-tuned FMG-Pan achieves competitive or even superior performance compared to the baseline, while significantly reducing runtime. On WV3, HQNR improves to 0.9644 with the lowest $D_\lambda$, and runtime drops to 1.47\,s per image. On WV2, HQNR remains comparable with a value equal to 0.9439, but the runtime decreases to 3.43\,s. These findings demonstrate that fine-tuning is a promising strategy to reduce training cost and accelerate deployment for our framework, without compromising (sometimes even improving) fusion quality.

\begin{table*}[!t]
\centering
\caption{Performance comparison between the baseline pretrained models and the proposed FMG-Pan framework on simulated data. The
reported values represent the average performance over 20 test images(\colorbox{red!10}{Red}: The improved results compared to the baseline.)}
\setlength{\tabcolsep}{4pt}
\renewcommand\arraystretch{1.25}
\resizebox{\linewidth}{!}{
\begin{tabular}{c|l|cccc|cccc}
\toprule
\multirow{2}{*}{\textbf{Dataset}} & \multirow{2}{*}{\textbf{Pretrained model}}
& \multicolumn{4}{c|}{\textbf{Baseline}}
& \multicolumn{4}{c}{\textbf{FMG-Pan}} \\
\cmidrule(lr){3-10}
 &  & SAM$\downarrow$ & ERGAS$\downarrow$ & SCC$\uparrow$ & Q2n$\uparrow$
    & SAM$\downarrow$ & ERGAS$\downarrow$ & SCC$\uparrow$ & Q2n$\uparrow$ \\
\midrule
\multirow{2}{*}{WV3}
& FusionMamba & 2.8245$\pm$0.5534 & 2.0890$\pm$0.5008 & 0.9874$\pm$0.0039 & 0.9212$\pm$0.0834
                 & {3.0803$\pm$0.6404} & {2.2188$\pm$0.5275} & {0.9849$\pm$0.0048} & {0.9129$\pm$0.0845} \\
& LAGNet       & 3.8258$\pm$0.7352 & 2.8503$\pm$0.7677 & 0.9738$\pm$0.0091 & 0.8722$\pm$0.1139
                 & \colorbox{red!10}{3.5971$\pm$0.7639} & \colorbox{red!10}{2.7380$\pm$0.6882} & \colorbox{red!10}{0.9757$\pm$0.0084} & \colorbox{red!10}{0.8971$\pm$0.0872} \\
\midrule
\multirow{2}{*}{WV2}
& FusionMamba & 5.5330$\pm$0.5352 & 4.2925$\pm$0.4078 & 0.9272$\pm$0.0078 & 0.8408$\pm$0.0888
                 & \colorbox{red!10}{5.3988$\pm$0.5930} & \colorbox{red!10}{4.1908$\pm$0.4356} & \colorbox{red!10}{0.9288$\pm$0.0073} & \colorbox{red!10}{0.8438$\pm$0.0882} \\
& LAGNet       & 8.6509$\pm$1.0960 & 7.3129$\pm$0.8368 & 0.7987$\pm$0.0452 & 0.6693$\pm$0.0941
                 & \colorbox{red!10}{5.9802$\pm$0.7432} & \colorbox{red!10}{5.0153$\pm$0.4367} & \colorbox{red!10}{0.8846$\pm$0.0175} & \colorbox{red!10}{0.8046$\pm$0.0801} \\
\midrule
\multirow{2}{*}{GF2}
& FusionMamba & 0.7076$\pm$0.1329 & 0.6206$\pm$0.1052 & 0.9923$\pm$0.0014 & 0.9830$\pm$0.0076
                 & {0.9437$\pm$0.1791} & {0.7955$\pm$0.1371} & {0.9868$\pm$0.0023} & {0.9696$\pm$0.0096} \\
& LAGNet       & 1.3639$\pm$0.2689 & 1.1957$\pm$0.2197 & 0.9672$\pm$0.0073 & 0.9376$\pm$0.0226
                 & \colorbox{red!10}{1.3217$\pm$0.2630} & \colorbox{red!10}{1.1931$\pm$0.2183} & {0.9667$\pm$0.0075} & \colorbox{red!10}{0.9416$\pm$0.0164} \\
\midrule
\multirow{2}{*}{QB}
& FusionMamba & 10.6492$\pm$0.7128 & 9.5755$\pm$0.8668 & 0.8622$\pm$0.0226 & 0.7182$\pm$0.1397
                 & \colorbox{red!10}{7.6485$\pm$1.7076} & \colorbox{red!10}{7.8914$\pm$0.8342} & \colorbox{red!10}{0.9015$\pm$0.0154} & \colorbox{red!10}{0.8178$\pm$0.0982} \\
& LAGNet       & 7.9303$\pm$1.5285 & 8.8132$\pm$1.0182 & 0.8721$\pm$0.0139 & 0.7506$\pm$0.0987
                 & \colorbox{red!10}{7.5542$\pm$1.5834} & \colorbox{red!10}{8.5561$\pm$1.0238} & \colorbox{red!10}{0.8805$\pm$0.0144} & \colorbox{red!10}{0.7674$\pm$0.0953} \\
\bottomrule
\end{tabular}
}
\label{tab:simulated_baseline}
\end{table*}

\section{Comparison with Test-time Fine-tuning}
\label{sec:cmp_ft}

Our method fundamentally differs from conventional fine-tuning paradigms in that it learns an external adaptive model under the guidance of a frozen pretrained model, rather than updating the parameters within the pretrained network itself. In contrast, traditional fine-tuning methods optimize weights inside the pretrained model, which requires maintaining a full computational graph during training. This results in substantial GPU memory consumption as well as increased computational overhead due to full-model forward and backward propagation.

To provide a comparison, we conduct experiments on the WV3 dataset using FusionMamba as the pretrained backbone. Both our method and the competing approaches are trained with the same spectral and physical loss functions. We compare against (1) full fine-tuning (Full FT) and (2) parameter-efficient fine-tuning (PEFT) using LoRA.

\begin{figure*}[!ht]
    \centering
    \includegraphics[width=1\linewidth]{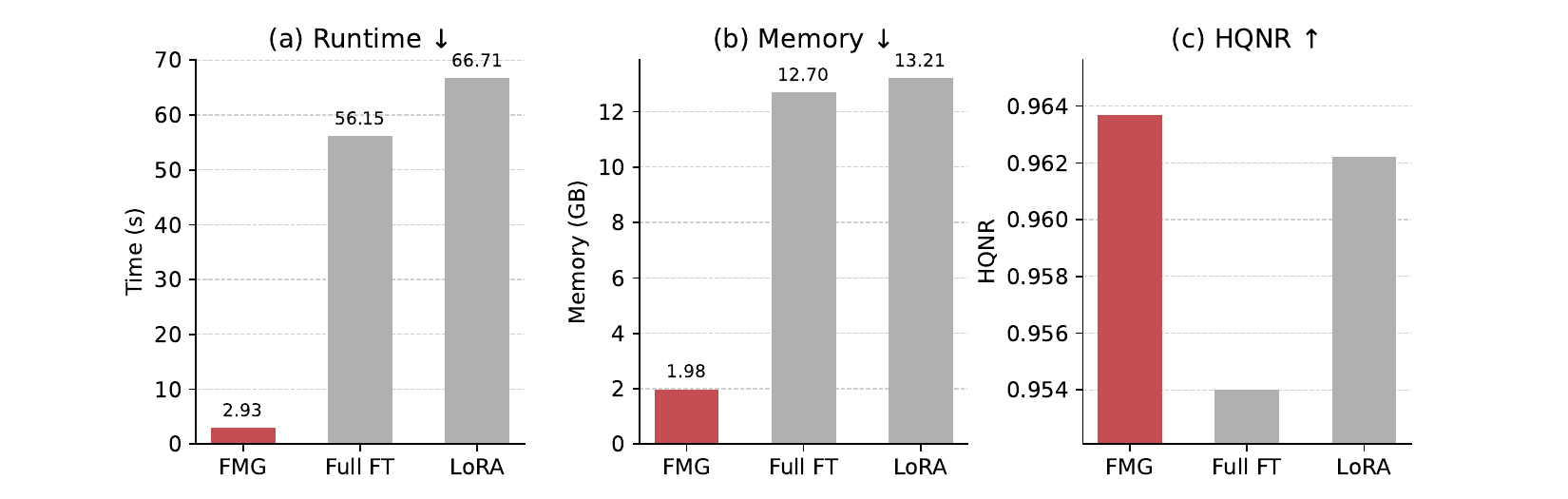}
    \caption{
\textbf{Comparison of efficiency and reconstruction quality on WV3.}
(a) Runtime, (b) GPU memory usage (c) HQNR.
FMG achieves substantially lower runtime and memory consumption compared to Full and LoRA, while maintaining competitive reconstruction quality.
}
    \Description{}
    \label{fig:CMPwithFT}
\end{figure*}

As illustrated in Fig.~\ref{fig:CMPwithFT}, the Full FT approach fails to converge to satisfactory performance even after 150 epochs. Although LoRA introduces only a small number of trainable parameters, it still relies on backpropagation through the large pretrained model, leading to a heavy computational graph and increased overhead. Consequently, LoRA exhibits slightly worse efficiency than Full FT in both time and memory consumption.

Moreover, both Full FT and LoRA are inherently constrained by the parameter space of the pretrained model. Even when optimized with our designed spatial and spectral losses, their performance remains inferior to our method. In contrast, our approach achieves the best quantitative metrics.
More importantly, FMG-Pan demonstrates significantly superior efficiency in both computational time and memory usage. This advantage becomes even more pronounced as the size of the pretrained model increases. Notably, once training begins, the memory consumption and training time of our method remain largely independent of the scale of the pretrained model, since no backpropagation is performed through it. These results collectively validate the effectiveness and practicality of our approach.

\section{Results on Simulated Data}
Although the primary goal of this work is to enhance the fusion quality and efficiency in real-world scenarios, we further conduct complementary experiments on simulated data from four datasets to provide a more comprehensive evaluation of the proposed framework. The pretrained strategies remain consistent with those used in the main experiments. As shown in Table~\ref{tab:simulated_baseline}, all pretrained models enhanced by FMG-Pan exhibit notable performance improvements under cross-sensor conditions, demonstrating the strong robustness of the proposed framework across different data distributions.

It is worth noting that for models such as \emph{FusionMamba}, which have been highly optimized on intra-sensor datasets (thanks to the abundant labeled data available for supervised training, making most metrics nearly saturated), introducing additional spectral or physical fidelity constraints may not yield further gains, and slight metric declines occur. This phenomenon only appears in the extreme configuration of “strongly supervised models with test data in the same-domain.” In contrast, for supervised models such as \emph{LAGNet}, which are not among the top-performing architectures, FMG-Pan still brings stable and practically meaningful improvements on intra-sensor datasets. These results indicate that the proposed single-image adaptation mechanism can effectively leverage pretrained priors and further enhance both the performance and generalization ability of pretrained models, even on simulated data.
\clearpage

\end{document}